\documentclass{jdmdh}
\usepackage{array}
\usepackage{pgfplots}

\usepackage{natbib}
\usepackage{graphicx}
\usepackage{url}

\title{Extracting Keywords from Open-Ended Business Survey Questions}
\author[1]{Barbara McGillivray}
\author[2]{Gard Jenset}
\author[3]{Dominik Heil}
\affil[1]{The Alan Turing Institute} 
\affil[1]{University of Cambridge}
\affil[2]{Independent Researcher} 
\affil[3]{University of the Witwatersrand, Johannesburg}

\corrauthor{Barbara McGillivray}{bmcgillivray@turing.ac.uk}

\begin{document}

%\received{20 March 1995; revised 30 September 1998}

%\pagerange{\pageref{firstpage}--\pageref{lastpage}}
%\pubyear{1998}

%\newcommand\eg{{\it e.g.\ }}
%\newcommand\etc{{\it etc}}

%\begin{document}

%\label{firstpage}
	\maketitle

% NOTE
% The preferred formatting system is LaTex, which can be used for direct typesetting, and a style file is available through https://mc.manuscriptcentral.com/societyimages/nle/NLE_LaTeX_Style_File.zip
%In case of difficulty, please contact cup-texsupport@aptaracorp.com

\abstract{Open-ended survey data constitute an important basis in research as well as for making business decisions. Collecting and manually analysing free-text survey data is generally more costly than collecting and analysing survey data consisting of answers to multiple-choice questions. Yet free-text data allow for new content to be expressed beyond predefined categories and are a very valuable source of new insights into people's opinions. At the same time, surveys always make ontological assumptions about the nature of the entities that are researched, and this has vital ethical consequences.
Human interpretations and opinions can only be properly ascertained in their richness using textual data sources; if these sources are analyzed appropriately, the essential linguistic nature of humans and social entities is safeguarded. Natural Language Processing (NLP) offers possibilities for meeting this ethical business challenge by automating the analysis of natural language and thus allowing for insightful investigations of human judgements.  We present a computational pipeline for analysing large amounts of responses to open-ended questions in surveys and extract keywords that appropriately represent people's opinions. This pipeline addresses the need to perform such tasks outside the scope of both commercial software and bespoke analysis, exceeds the performance to state-of-the-art systems, and performs this task in a transparent way that allows for scrutinising and exposing potential biases in the analysis. %\cite{church2017emerging}. %I've taken this out because I don't think it's good to have references in an abstract, and an abstract should stand alone.
Following the principle of Open Data Science, our code is open-source and generalizable to other datasets.}


\section{Context and motivation }

Leaders, managers, and decision-makers critically rely on information and feedback. Decision-makers first need information about the current set of circumstances which provide the context of the decision, and then need feedback on how the decision could play out. To get such information in a format that allows them to appropriately understand the entity they are seeking to comprehend is of critical importance to come to a high-quality decision. Often only qualitative insight into the opinions, interpretations and assumptions of large numbers of people will allow us to understand a set of circumstances properly and are therefore required to make high-quality decisions and consequently outcomes.
In the context of surveys, qualitative or open-ended answers are valuable since on the one side they acknowledge humans as the kinds of beings who interpret and re-interpret the world; on the other hand, by expressing precisely this human quality, surveys allow respondents to express unique perspectives and nuances that the researchers may not have been able to foresee. However, researchers have often reached for quantitative or closed-form questions where answers are expressed on a numerical scale, since this means the respondent is the chief source of the error of measurement \cite[104]{gorard_2003}. The relative ease with which numerical answers can be analyzed compared to open-ended answers is an obvious explanation for this preference. However, research inevitably makes ontological assumptions and claims about the nature of the object of study. For instance, \citet[104]{gorard_2003} gives the example of reported crime, where reported crime levels were higher when the survey instruments used closed scales, compared to open-ended questions. Furthermore, there is evidence suggesting that respondents are more consistent when using a self-defined or open-ended numerical rating scale, compared to a closed numerical scale \cite[218]{johnson_2008}. Thus, in addition to the extrinsic value of open-ended survey questions (e.g.\ in the form of improved revenue or business decision making), there are also clear ethical implications stemming from the choice of research instrument. We argue that the practical obstacles to analyzing open-ended answers should be met by an increased focus on open-source Natural Language Processing (NLP) applications that focus on aiding understanding. Such a move inevitably requires merging the linguistic facts with domain knowledge \cite[766]{friedman2013natural}.

In suggesting research based on qualitative data we are confronted with a challenge. Research on quantitative data is naturally well-suited for analysing large amounts of information. On the other hand, the analysis of free-text responses has been notoriously tedious and time-consuming as it up to now typically required significant human intervention. In addition to this, there is the constant challenge of analysing such free-text data in an unbiased way, where human judgement is minimized in favour of an objective description and transparent analysis of the data. The risk of confirmation bias when working with textual and linguistic data was pointed out already by \citet[97]{kroeber1937quantitative}, noting that it is all too easy to unconsciously overlook evidence opposing some interesting or desirable affiliation in the data, while mentally taking note of every item corroborating it.

The particular challenge that the proposed analysis seeks to address is that on the one hand the sample sizes are too big to be analysed manually. On the other hand, 
%the sample sizes are too small, leading to unacceptably large margins of error in the categorisation of individual statements. 
error margins are tolerated in analyses of large amounts of text, such as Twitter feeds, because they are likely to be evened out due to the large numbers of statements analysed, but the sample sizes at hand are too small allow for such large margins of errors. This dilemma -- the availability of too much text to analyze manually, and the need for small error margins 
%reliably analyze in a statistics-driven fashion 
-- is at the heart of the argument made in this paper.

In this context, NLP provides an opportunity to develop solutions that allow increasing levels of automation in processing and analysing qualitative data input. Automating the processing of qualitative data also allows us to analyse and address issues of bias in a more structured way than would be the case when the process is manual. The inherent transparency of the approach proposed here makes it easier to analyse the elements of the linguistic algorithm that generates a bias compared to doing the same for a human being who generates a bias. This permits a more structured and transparent interrogation to explore which parts of the approach cause a bias and find technical solutions to reduce such bias.

In this article we focus on a limited linguistic domain, namely the set of employee responses to a specific survey question. Our aim is to apply a series of NLP and Information Retrieval techniques with the practical goal of developing an open-source pipeline for extracting keywords from free-text survey responses that can be employed by non-specialists. The applied nature of this work aims to fill the gap between the algorithms developed in the context of research in theoretical computational linguistics and commercial software that provides so-called ``black-box'' systems that cannot be easily modified and adapted.

\section{Philosophical and ethical argument, and example}

Typically, research is commissioned to help the decision-making process in order to chart an appropriate course of action to address an issue or multiple issues. Because such research is intended to inform action, the assessment of the appropriateness of such action falls into the field of ethics. Ethics, broadly speaking, is the inquiry of whether the decisions made or actions taken based on research are desirable or undesirable, better or worse, right or wrong. Therefore, any technology that can improve the insight gained from such research has positive ethical import.

For research to support desirable outcomes, it needs to be not only based on a methodology which is applied correctly and thus leads to correct outcomes, but it also needs to emphasize those aspects of the entity or entities to be researched that are relevant to making a desirable decision of leading constructive action. Such research needs to not only apply a given methodology correctly, but also needs to apply the methodology that gives a good understanding of the entity or entities to be researched. To get to an understanding of the aspects of an entity or a group of similar entities is the domain of ontology. Ontology in its broad definition is understood as the branch of metaphysics dealing with the nature of being and is also understood to be the study of beings as such, because we need an understanding of what something is in the first instance before we can decide what is desirable or undesirable for an entity.

Intuitively we know that certain aspects of human life, such as true friendship, can hardly be quantified, and in some cases quantification can be downright misleading, for example when evaluating art by the price it can command in an auction. It should therefore already be clear that quantifying certain phenomena in itself has an ethical import, whether this ethical import is explicitly understood or not in each case when it is applied.

As \citet{morgan1980case} pointed out some time ago, in committing to a particular research methodology, the researcher makes particular assumptions about the entity to be researched, whether the researcher is aware of this or not. Morgan and Smircich point out that choosing quantitative methods for researching social phenomena, one makes the assumption that the social world consists fundamentally of concrete structures similar to those entities researched in the natural sciences \cite[498]{morgan1980case}:

``Once one relaxes the ontological assumption that the world is a concrete structure, and admits that human beings, far from merely responding to the social world, may actively contribute to its creation, the dominant [quantitative] methods become increasingly unsatisfactory, and indeed, inappropriate.''

And they go further ``to point to a neglected feature of \textit{all} social research -- that it is based on implicit and untested ground assumptions.'' \cite[499]{morgan1980case}. The inquiry of testing ground assumptions falls into the domain of ontology, which interrogates the very nature of entities such as people and human collectives. In particular, in the twentieth century philosophers like \citet{wittgenstein1953philosophical}  and \citet{heidegger}  demonstrated that core to humans is their relation to language and their ability to interpret and re-interpret the world both individually and collectively. It can be deduced from there that any research that does not articulate such interpretations does not capture what is essential to humans and human communities. It should therefore be obvious that only research that acknowledges the qualitative nature of the data can achieve this.

It is noteworthy that these comments by Morgan and Smircich were made already in 1980, in presumably one of the most prominent publications for management research, arguing for the adoption of more qualitative research for understanding social phenomena. Presumably the reason why qualitative research has not gained more prominence in research related to social phenomena is twofold. The first one is articulated by \citet{powell_2003} that the dominant approaches of research are not open to be investigated by the required ontological discourse. This is done without noticing that related research approaches are actually not ontologically neutral and lead to ethically questionable and misleading research \cite[27--32]{heil2011ontological}.

It is beyond the scope of this paper to give a comprehensive overview of an appropriate ontological investigation into social phenomena and their ethical implications, and suggest specifics about appropriate research approaches beyond stating that qualitative research should play a much more prominent role in the research of social phenomena. Rather, this paper will focus on the much more practical question of why qualitative research has not become more prominent. This reason is that doing qualitative research on a large scale so far has been prohibitively time-consuming. It is hardly fruitful to criticize quantitative research methodologies for their ontological inadequacies when at the same time it is simply not feasible to do qualitative research in such a way that it can be applied on a large scale and with high levels of confidence to give a representative understanding of a large population. Rather, we hope that by contributing to making qualitative research significantly less time-consuming, the insights gained from such research will speak for themselves, and this supports the ethical and ontological claims by demonstrating them in practice. As such, the approach we are putting forward is intended to be a prime example of a technology with positive ethical import at the most fundamental level.

\section{Background}

As mentioned above, free-format text data are valuable, but unlike structured data such as ratings or binary questions, they are not easily accessible for analysis without substantial pre-processing. Such pre-processing can be done manually, but the process is time-consuming and does not scale well to large sets of data. The alternative to such manual analyses is to turn to computational techniques from the domains of NLP and information retrieval.

\subsection{NLP algorithms for keyword extraction}

Keyword extraction remains an unsolved problem with lower performance than in many other NLP tasks \citep{hasan2014automatic,lahiri2017keyword}. Keyword extraction is used both in general domains, such as news articles \citep{hasan2014automatic}, and more narrow genres such as medical texts or scientific writing \citep{hasan2014automatic, friedman2013natural}. The approach can also be based on specific channels of communication, such as emails \citep{lahiri2017keyword} or Twitter messages \citep{abilhoa2014keyword}. To our knowledge, no work has been done specifically targeting keyword extraction for survey response data. The general difficulty of the keyword extraction task depends on several factors, including the length of the text, their structural consistency, the number of topics in the texts and how they correlate \citep{hasan2014automatic}.

Typically, keyword extraction is a multi-step process that starts with basic text processing to ensure higher quality input to the keyword candidate steps \citep{lahiri2017keyword}. An initial step is tokenization, the process of splitting the text into individual discrete units (tokens) for further analysis. Building on tokenization, part of speech tagging assigns a part of speech to each word or token in the text. Further, lemmatization makes use of the output from the two previous algorithms for grouping tokens together based on their lemma, or dictionary look-up form. Such pre-processing steps are useful to obtain an intermediate-level overview of free-form text, since they enable easy searches for all forms of a lemma, e.g. \textit{buy} for  `bought', `buying', `buys', or all words with a given part of speech, e.g. all nouns in a dataset. However, such pre-processing is not sufficient to convey a high-level semantic overview of the data at hand. Following Zipf's law, which states a statistical relationship between the logarithmic frequency of a word and its logarithmic rank in a list of ordered frequencies \cite[15]{baayen2001word}, the most frequent words in a text, displaying the highest Zipf ranks, tend to be function words \cite[17]{baayen2001word}, which offer little insight on the content of a text, and are typically removed.

Further algorithms have been developed that build on pre-processed data, aiming to extract the most salient keywords from free text data, thus providing a keyword summary of the text. Automatic keyword extraction at scale can be done by means of linguistic rules, statistical approaches, or machine learning approaches. See \cite{siddiqi2015keyword} for an overview. 
To some extent, the methodology is field dependent. In corpus linguistics, a field of research that relies heavily on statistical metrics,
statistical approaches are widely used for studying keyword-related phenomena \citep{pojanapunya2018log}. Other approaches focus on 
the vector space representation of the words themselves, rather than on formal statistical tests \citep{hu2018domain}. Graph based approaches, among them RAKE \citep{rose2010automatic},
are widely used for automatic keyword extraction, in several adapted variants \citep{zhang2018adapted}.  A related effort involves grouping words or keywords together into clusters or topics. Such topic extraction represents an added level of processing for dealing with synonyms such as \emph{staff} and \emph{employee}, or word senses such as \emph{screen} in the sense of physical display and \emph{screen} in the sense of what a software application displays. In this article we will focus on keyword extraction.

The crucial step in any keyword extraction pipeline is the filtering stage, where potential keywords are ranked and evaluated. The means to perform this varies, and the approaches can be categorized in different ways \citep{hasan2014automatic,beliga2015overview}. From a machine learning point of view, the problem of keyword extraction can be recast as a supervised or unsupervised classification problem \citep{hasan2014automatic}. A deeper focus on the methods to achieve the classification reveals a wider range of methods \citep{beliga2015overview}. These include simple distributional methods such as TF-IDF (Term Frequency-Inverse Document Frequency), linguistic approaches that use lexical, syntactic, and semantic features, (supervised) machine learning, (unsupervised) vector space models, and graph models that model texts as graphs whose nodes are words and whose edges are relations between words, e.g.\ co-occurrence relations. These approaches are not mutually exclusive, so that linguistic features can be used as training features in a supervised classifier, for instance.

Evaluation of keyword extraction requires a manually annotated gold standard. Typical evaluation measures include precision, recall, F-scores, and the Jaccard index \citep{hasan2014automatic,friedman2013natural,lahiri2017keyword}. Since keywords are contextual, a high performance in one channel or domain cannot be taken as absolute indication of high performance in another. Equally, some domains or channels might be more challenging than others, according to the dimensions mentioned above. This situation opens an interesting scenario when it comes to practical applications of keyword extraction. Without a uniform task difficulty, some domains or channels might be more susceptible than others to certain approaches. Since some keyword extraction algorithms are more challenging to implement than others, this implies that the threshold for adopting keyword extraction successfully in a domain or channel might vary substantially. Such a threshold for adoption will depend on several factors, not least what is already available out of the box. The subsequent section provides a brief overview of some existing packages and frameworks, with an eye towards their application in keyword extraction for survey data.

\subsection{Existing software packages}

A number of software packages exist for processing free-form text data. A full overview falls outside the scope of this article, but packages such as RapidMiner and IBM's SPSS Text Analytics Modeler provide keywords and topics. Additionally, a number of packages aimed at the corpus linguistic research community exist, such as WordSmith, MonoConc, and AntConc, and have a rather narrow, specialized focus.
In addition to the software packages listed above, a number of programming languages implement toolkits for processing text. Interpreted languages such as Perl and Python have a long history in text processing, and in the case of the latter the NLTK toolkit has a wealth of NLP functions available. Even a specialized language such as the R platform for statistical computing has user-contributed packages for text mining, e.g.\  \citet{meyer2008text}. For a non-scripting language such as Java, toolkits include OpenNLP, Stanford NLP, LingPipe and GATE, to mention a few. The GitHub keyword extraction topic page provides an overview of some of code available, in several different programming languages (\url{https://github.com/topics/keyword-extraction}, accessed 14/04/2019). As with the other packages mentioned above, a full survey is beyond the scope of the present article.

Whereas the software packages mentioned above lack flexibility, the programming language toolkits have an abundance of flexibility, but typically require a deep technical expertise to be used. Additionally, since software packages tend to come with a graphical user interface, they require extra (external) documentation. Documentation is crucial when working with free text data, because, as noted above, the process of extracting information from such data nearly always involves a series of steps that follow one another in a sequence. Since each step is dependent on the output of the previous step, fine-tuning the whole process might require manipulation of parameters and settings in several software components. In an iterative process, or a process being applied to a new dataset, documentation is typically required to have full control over, and thus a complete understanding of, the final information being produced. Conversely, computer code lends itself well to reproducible, documented multi-module processes, but the learning curve and the time investment might prove prohibitive.

In short, while software packages tend to be less flexible (as well as sometimes costly), programming languages and their toolkits are flexible but complex, requiring both expertise and development time. We believe that a reasonable compromise is an open-source pipeline implemented in a high-level language such as Python. We prefer Python over Java or C++ because we have observed that Python is easy to learn for practitioners or academics with a non-computer science background, making it easier for them to contribute to existing code. Moreover, Python has a large community of programmers, which provides valuable support, as well as a wide range of libraries for NLP. Additionally, Python packages PySurvey (\url{https://pypi.python.org/pypi/PySurvey/0.1.2}, accessed 14/04/2019) already exist to deal with numerical survey data. Extending Python's survey capabilities to cover free text responses seems a natural next step.

\section{The pipeline}

A modular pipeline approach is desirable, since it allows different modules with different algorithms to feed into each other, thus producing a reasonably unbiased, objective summary of the input text. However, a knowledge extraction process can never be completely neutral or objective. For this reason, we stress the need for open source NLP systems and modules, where the assumptions of the developers can be scrutinized in detail, even at the implementation level. This alone constitutes an advantage for open source systems over proprietary ones, such as SPSS's Survey Analyzer. Such a philosophy has the drawback that it risks separating those with the programming skills to both use and evaluate existing code or modules, from those without such skills, who have no choice but to rely on proprietary software. Furthermore, an open toolkit allows further developments and improvements to be shared. We argue that a desirable pipeline should have the following properties:

\begin{itemize}
\item Extensive reuse of existing modules where possible;
\item Focus on relatively user-friendly pipelines, where the focus is on outputs that aid understanding of the data;
\item Open-access code, in a high-level language such as Python;
\item Practical usability combined with transparency.
\end{itemize}

This section will demonstrate how the pipeline we have developed fulfils these criteria. We implemented the pipeline using the open-source programming language Python, version 2.7 and we are in the process of creating a version that is compatible with Python's version 3. The code is available on GitHub at the url \url{https://github.com/BarbaraMcG/survey-mining}.

\subsection{Pipeline components}

The pipeline consists of the following main components, organized in a modular way:
\begin{enumerate}
\item Initialization
\item Text pre-processing
\item Keyword extraction
\end{enumerate}

The input to the code is a file (in a spreadsheet format) which contains one row per survey response. In the test dataset we have used, for example, the following survey question was submitted to the employees of a private telecommunication company: ``What makes you proud of working in this company?'' and the input file was an Excel spreadsheet with 599 responses. Table \ref{table-examples-responses} shows some of the responses from this dataset.

The code also takes as input a file containing a tailor-made list of acronyms, compiled specifically while keeping in mind the context and sector in which the company operates, and contains acronyms such as \emph{IT} and \emph{ICT}. This list is used in the code to exclude such acronyms from the list of keywords it produces, but is not a strict requirement for the pipeline.

In the initialization stage, the code collects the names of relevant input and output directories and files; it also adds the target word, i.e.\ the object of the survey question (\emph{pride}, in the example above), and the company's name to the set of words to be excluded from the keyword list, which is pre-populated with the acronyms contained in the input file.

In the text pre-processing phase, line breaks, bulleted lists, and numbered lists are deleted from the response texts. Next, the response texts are tokenized, part-of-speech tagged, and lemmatized using the tokenizer, part-of-speech tagger, and the WordNet lemmatizer from the NLTK package \citep{nltk_book}.

The initial keyword extraction is performed using the Python library topia.termextract \citep{topia}. In addition, we extracted keyword noun compounds when two noun keywords occurred one after the other; an example of this is /emph{training centre} in the third response from Table \ref{table-responses-kws-pipeline}. In order to add a further level of analysis, we also extracted any adjectives modifying the keywords, together with their frequencies. This heuristic has the advantage of bringing about a linguistically richer set of keywords. As in \citet{lahiri2017keyword}, the term ``keyword'' is used collectively about words and phrases.

The output of the pipeline consists of two files. The first file displays, for each response, the list of its keywords and their modifying adjectives. As an example, see Table \ref{table-responses-kws-pipeline} for the output relative to the third response from Table \ref{table-examples-responses}. The second file lists all keywords extracted from the dataset, with their frequency (i.e.\ the number of responses containing it), and the adjectives modifying them, with their frequency (i.e.\ the number of responses containing the keyword and the adjectives together).

%\begin{table}
 % \centering
  %\begin{tabular}{+>{\bfseries}l^c^c^c^c}
   % \hline
    %\rowstyle{\bfseries}
    %& Sepal.Length & Sepal.Width & Petal.Length & Petal.Width\\
    %Setosa & 5.006 & 3.428 & 1.462 & 0.246\\
    %Versicolor & 5.936 & 2.77  & 4.26  & 1.326\\
    %Verginica & 6.588 & 2.974 & 5.552 & 2.026\\
    %\hline
 % \end{tabular}
%  \caption{Morbi malesuada diam at magna condimentum.}
 % \label{tab:example}
%\end{table}

% This is table 1
\begin{table}
  %\newcolumntype{+}{>{\global\let\currentrowstyle\relax}}
  %\newcolumntype{^}{>{\currentrowstyle}}
 % \newcommand{\rowstyle}[1]{\gdef\currentrowstyle{#1}%
 %   #1\ignorespaces
 % }
    \begin{tabular}{l}
    \hline
%    \rowstyle{\bfseries}
    Response\\
    \hline
    Providing services e.g.\ fibre to the home \\ 
    \parbox[t]{7cm}{Get work satisfaction working with new \\technologies +constantly learning}\\
    You can further your studies, have an internal training centre \\
    \parbox[t]{7cm}{Where we've come from challenges we have \\faced +now giving investors returns}
   \\ 
    \hline
    \end{tabular}
    %\vspace{-2\baselineskip}
    \caption{Example survey responses describing what makes the respondent proud to work for the company.}
  \label{table-examples-responses}
\end{table}

% This is table 2
\begin{table}
  
    \begin{tabular}{lcc}
    \hline\hline
    Response & Keyword & Adjectives \\ \hline
    You can further your studies, have an& study & \\internal training centre. && \\\hline
    You can further your studies, have an & training centre & internal \\internal training centre.&& \\\hline
    You can further your studies, have an& training & internal \\internal training centre. &&  \\\hline
    You can further your studies, have an& centre & \\internal training centre. && \\
    \hline\hline
    \end{tabular}
    %\vspace{-2\baselineskip}
    \caption{Subset of the keywords extracted by the pipeline (second column), together with the adjectives modifying them (third column), and the corresponding survey response (first column, corresponding to the third response from Table \ref{table-examples-responses}).}
  \label{table-responses-kws-pipeline}
\end{table}

\section{Evaluation}

To our knowledge, there is no gold standard dataset for keyword extraction evaluation. To evaluate our pipeline, we compared it with a manually annotated dataset. Furthermore, since our work sits at the interface between research and NLP applied to commercial contexts where the tools can be used by non-specialists, we also used the output from a commercial software package specifically aimed at survey data. Finally, we also compared the pipeline against a baseline open computational method (TF-IDF).

In order to prepare a comparison set for the evaluation, we manually analyzed the test dataset. This analysis aimed at a high-level description of the content of each survey response by means of keywords; an example of such keywords is given in Table \ref{table-responses-kws-manual} and relates to the second response in Table \ref{table-examples-responses}.

% This is table 3
\begin{table}
  
    \begin{tabular}{lrr}
    \hline\hline
    Response & Keyword & Description \\ \hline
    \parbox[t]{3cm}{Get work satisfaction\\constantly learning\\ working with new\\ technologies + constantly learning} &new technology &  \parbox[t]{3cm}{New technology\\ and innovation} \\\hline
    \parbox[t]{3cm}{Get work satisfaction\\ working with new\\ technologies + \\constantly learning} & learning & \parbox[t]{3cm}{Skills development, learning,\\ personal development} \\\hline
    \parbox[t]{3cm}{You can further your\\ studies , have \\a internal \\training centre.} & training centre & \parbox[t]{3cm}{Skills development, learning,\\ personal development}  \\\hline
    \parbox[t]{3cm}{Give better security\\ re new products \\launched.} & new product &  \parbox[t]{3cm}{New technology\\ and innovation} \\
    \hline\hline
    \end{tabular}
    %\vspace{-2\baselineskip}
    \caption{Subset of keywords manually assigned to the second response from Table \ref{table-examples-responses}}
  \label{table-responses-kws-manual}
\end{table}

As we can see from Table \ref{table-responses-kws-manual}, the keyword assigned to the first response text is ``new technology''; the adjective \emph{new} associated with technology is important to convey the fact that one of the company's qualities is its innovation. This example demonstrates the need to extract adjectives of the keywords, in addition to the keywords themselves.

For the comparison with commercial software, we chose SPSS Survey Analyzer 4.0.1 because it is specifically designed to handle survey data. Additionally, following \citet{liu2009unsupervised}, the baseline method we chose was keyword extraction with TF-IDF. TF-IDF  is a well-known metric for statistically retrieving topic or index terms associated with a collection of documents \citep{sparck1972statistical}. TF-IDF is both simple to compute and widely used for practical applications \citep{sparck2004idf}. In order to demonstrate added value, our new pipeline ought to show improved results over a simple TF-IDF approach.

When comparing the set of keywords extracted by our system with the baseline, the set from the manual analysis, and the one from SPSS, we performed exact match after stemming, i.e.\ we first stemmed the keywords from all sets, and then assessed when the stemmed keywords appeared in the different sets. Table \ref{table-kws-nb} displays the number of stemmed keyword types in each system.

The purpose of the evaluation was to assess the quality of the pipeline against the other methods (commercial software, baseline, and manual analysis) according to the following metrics:

\begin{enumerate}
    \item Response coverage: number of responses for which at least a keyword was found in each of the four systems;
    \item Precision, recall, and F-score for each response, calculated by comparing each system with the manual analysis;
    \item Correlation between the set of frequency-ordered keywords produced by the manual analysis and the set of keywords produced by each of the three systems;
    \item Similarity between the set of keywords produced by the manual analysis and the set of keywords from the three systems, on a per-response basis.
\end{enumerate}

% This is table 4
\begin{table}
  
    \begin{tabular}{ll}
    \hline\hline
    Data set & Number of stemmed keyword types \\
    \hline
    Manual analysis & 1,163\\
    Pipeline & 365\\
    SPSS & 37\\
    Baseline (TF-IDF) & 471\\
\hline\hline
    \end{tabular}
    %\vspace{-2\baselineskip}
    \caption{Number of stemmed keyword types in each of the systems used in the evaluation.}
  \label{table-kws-nb}
\end{table}

Table \ref{table-kws-nb} displays the number of keywords and Table \ref{table-coverage} displays the coverage metric, for the four approaches: manual analysis, the pipeline, SPSS, and the baseline keyword extraction with TF-IDF.
We note that SPSS's coverage is the lowest one. This can be explained by the fact that this software package extracts a low number of keywords compared to the other methods considered. Furthermore, note that the manual analysis has a lower coverage compared to the pipeline and the baseline.

The first row in table \ref{table-evaluation} displays the number of keywords shared by the set extracted by the manual analysis on the one side and the set extracted by the pipeline (first column), SPSS (second column), and the TF-IDF baseline (third column) on the other side. If we consider the manually extracted keywords as a gold standard, this represents the absolute number of correct keywords in each of the three systems. The second row shows the precision of each of the three systems, defined as the number of correct keywords out of all extracted keywords. The third row shows the systems' recall, defined as the number of correct keywords out of all keywords from the manual analysis. Finally, the fourth column contains the F-score, defined as \[F1 = 2*\frac{precision*recall}{precision+recall}.\] As we can see from Table \ref{table-evaluation}, the pipeline outperforms (or equals) the other two systems in all three metrics.

% This is table 5
\begin{table}
  
    \begin{tabular}{ll}
    \hline\hline
    Data set & \% Coverage\\
    \hline
     Manual analysis & 93\\
     Pipeline & 100\\
     SPSS & 49\\
     Baseline (TF-IDF) & 100\\
 \hline\hline
    \end{tabular}
    %\vspace{-2\baselineskip}
    \caption{Coverage of each system in the evaluation.}
  \label{table-coverage}
\end{table}

Another criterion for evaluating the pipeline concerns how accurately it describes the frequency of the keywords in the dataset, i.e. the number of responses containing each of the keywords. Table \ref{table-freq} shows the top 10 keywords extracted by the pipeline, with their frequencies.

% This is table 6
\begin{table}
  
    \begin{tabular}{llll}
    \hline\hline
    Measure & Pipeline & SPSS & Baseline (TF-IDF)\\
    \hline
    Number of correct keywords  & 152 & 31 & 138\\
    Precision & 0.33 & 0.33 & 0.26 \\
    Recall & 0.42 & 0.09 & 0.38 \\
    F-score & 0.37 & 0.14 & 0.31 \\\hline\hline
    \end{tabular}
    %\vspace{-2\baselineskip}
    \caption{Different evaluation metrics for the three systems analyzed.}
  \label{table-evaluation}
\end{table}

We compared the list of keywords frequencies of the manual analysis against the three methods and ran Spearman's correlation test \citep{daniel1990spearman}. This test assesses monotonic relationships between two datasets. Unlike Pearson's correlation test, this test does not assume that both datasets are normally distributed. The Spearman's correlation coefficient varies between -1 and +1; a coefficient equal to 0 shows no correlation; if the coefficient is positive and close to 1, this means that, as one dataset increases, so does the other one and that this association is strong. The results of the correlation tests are shown in Table \ref{table-corr}.

% This is table 7
\begin{table}
  
    \begin{tabular}{ll}
    \hline\hline
    Keyword & Num.\ responses\\\hline
    company & 104\\
    service  & 83\\
    product & 75\\
    work & 65\\
    benefit & 56\\
    offer  & 49\\
    customer & 45\\
    job  & 40\\
    technology  & 38\\
    staff  & 36\\\hline\hline
    \end{tabular}\caption{Number of survey responses containing the top 10 keywords at least once. Note that each response may contain more than one keyword.}
    %\vspace{-2\baselineskip}
  \label{table-freq}
\end{table}

The test gave significant results in all cases, and showed a moderate correlation between the manual analysis and the pipeline and between the TF-IDF baseline and the manual analysis, and strong correlation between the SPSS and the manual analysis. According to this measure, the pipeline performs worse than the other two systems. However, a correlation is always calculated for vectors of equal length. This means that correlation could only be calculated for the overlap between the gold standard and each of the sets of extracted keywords. Since SPSS only identified a very small number of keywords, the correlation coefficient is based on a much smaller sample compared to the pipeline or the TF-IDF baseline.

Finally, for each response, we compared the set of stemmed keywords extracted by each system again the manual analysis, in order to account for the general distribution of the similarity between the systems.
We calculated the Jaccard coefficient \cite[299]{manning1999foundations} to measure the similarity between the keyword sets. This coefficient tends to penalize small number of shared entries, which is appropriate to the case at hand. %An example is given in Table \ref{table-jaccard}.

% This is table 8
\begin{table}
  
    \begin{tabular}{lll}
    \hline\hline
    Dataset & Spearman's correlation & $p$-value\\ \hline
    Pipeline & 0.30 & $<<0.05$ \\
    SPPS & 0.73 & $<<0.05$ \\
    Baseline (TF-IDF) & 0.44 & $<<0.05$ \\
   \hline\hline
    \end{tabular}
    %\vspace{-2\baselineskip}
    \caption{Results of correlation test on the  list  of  keywords  frequencies  of  the  manual  analysis  against the  three  methods in the evaluation.}
  \label{table-corr}
\end{table}

\begin{figure}
  % Requires \usepackage{graphicx}
  \includegraphics[width=0.7\textwidth]{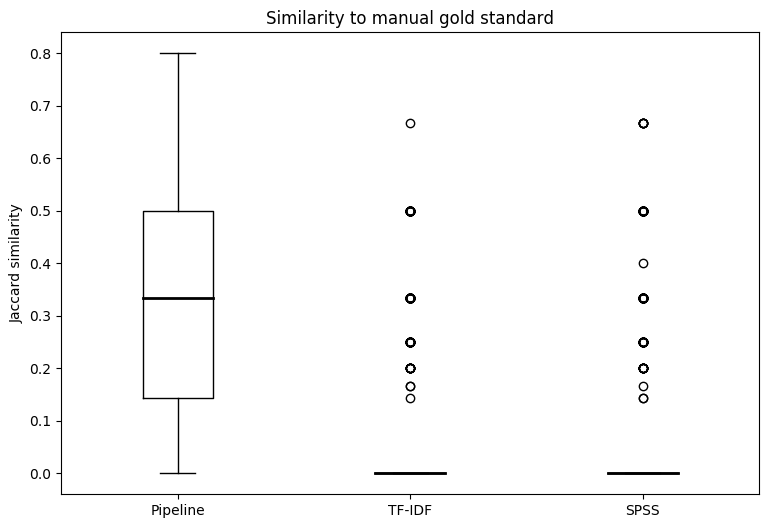}\\
  \caption{Box and whiskers plot comparing the Jaccard similarities for the Python pipeline, SPSS, and TF-IDF.}\label{fig_jaccard}
\end{figure}

Figure \ref{fig_jaccard} shows a box and whiskers plot of the Jaccard similarity between the manual analysis and the three methods and Table \ref{table-jaccard} shows the summary statistics of the distributions in terms of minimum, maximum, mean, and standard deviation.

As we can see from Table \ref{table-jaccard}, our pipeline's performance is the best one, because the keywords it extracts are closer to the manually-extracted keywords compared to the TF-IDF and SPSS systems.

% This is table 9
\begin{table}
  
    \begin{tabular}{llll}
    \hline\hline
    Jaccard Index & Pipeline & TF-IDF & SPSS \\
    \hline
    Minimum & 0.00 & 0.00 & 0.00 \\
    Maximum & 0.80 & 0.70 & 0.70 \\
    Mean & 0.30 & 0.09 & 0.09 \\
    Standard deviation & 0.22 & 0.16 & 0.17 \\
    \hline
    \end{tabular}
    %\vspace{-2\baselineskip}
    \caption{Summary statistics of the distributions of Jaccard similarity between the keywords extracted with manual analysis and the keywords extracted by the three methods.}
  \label{table-jaccard}
\end{table}

\section{Discussion}

The pipeline discussed above seeks to fill a gap in the field of keyword extraction by focusing explicitly on free-text survey response data. As mentioned above, some commercial software packages with such a focus exist, in addition to general approaches such as TF-IDF. However, the special nature of survey responses, namely their short length and dependence on the context provided by the survey question, begs the question of how well a general approach will fare. As the evaluation of the pipeline in section 5 demonstrated, TF-IDF performed quite well, but still worse than our pipeline. This might be explained by the fact that, as a result of having fewer words, short texts will also have a lower TF-IDF score. This will penalize terms during the extraction phase not due to the frequencies of the words themselves, but due to the length of texts they appear in. The commercial SPSS add-on package covered a small number of the same keywords as the pipeline, but also missed out on a high proportion of the keywords in the manually-analyzed data and overall performed worse than our pipeline in all but one evaluation metric, the one based on Spearman's correlation.

In contrast, our linguistically motivated unsupervised pipeline performed well. Compared to the gold standard, the pipeline achieved an F-score of 0.37. This result is comparable with the F-scores reported by \citet{lahiri2017keyword}, who report a maximum F-score of 0.39 for a keyword extraction system for email data. We also saw that the Jaccard similarity for the pipeline keywords was much higher than for the other approaches.  This result may indicate that short survey answers, with their crucial dependence on the semantic and pragmatic coherence provided by the context of the survey question, benefit particularly from the pipeline's use of linguistic information.

Nevertheless, we should consider one possible objection and limitation to the pipeline, the question of scalability. The current pipeline was not constructed with Big Data technologies in mind, i.e. large-scale distributed systems for dealing with datasets too large to be processed on a single computer. However, we do not consider this a major objection, given the intended use case. Although some surveys might collect  data truly too large for a single computer to process, there are probably more survey data collected on a smaller scale, sufficiently small to be processed on a single computer. In this respect the pipeline is similar to a commercial single-computer application for mining survey data. However, due to the open and transparent nature of the pipeline, it can be re-implemented for a Big Data system, should the need arise. This flexibility constitutes an advantage over proprietary, commercial systems.
Furthermore, when it comes to free-form text data, even a small or medium-sized sample, such as the one analyzed here, is sufficiently large to present the researcher with complexities of a sufficient scale that automated, unsupervised processes are valuable.

The interpretation process itself, which is never entirely free from human judgment, can be improved by the addition of an automated step. This is not because such a step is closer to some pre-existing objective reality, but simply because it performs predictably and consistently in its assigned tasks. In addition to the sheer workload of manually analyzing free-format text survey responses, a researcher resorting to intuition alone might easily become biased when noting ``a certain affiliation that is real enough, but perhaps secondary; thereafter [noting] mentally every corroborative item, but unconsciously overlooks or weighs more lightly items which point in other directions.'' \cite[97]{kroeber1937quantitative}. An automatic keyword extraction process helps protect against bias and unconscious oversights, while offering valuable input for the human interpretation performed by the researcher.

The choice of keyword extraction system will always depend on a number of variables. We have shown that our pipeline stands up well compared with some alternative approaches. Beyond the performance in terms of specific metrics, it is worth noting the arguments by \citet{church2017emerging} and \citet{manning2015computational} regarding the importance of transparency in computational NLP models. The same point can be extended to analysis of survey data. A linguistic or hybrid keyword extraction approach will by its nature be more transparent than a purely statistical approach, other things being equal. Such transparency is not only of interest in academic research, but also in the realm of applied NLP for decision making, where the transparency supports accountability.

Another consideration is the nature of the text to be analyzed. In our case, the application area was survey questions, which have a number of special features. In most cases they are short, and since short texts will yield fewer potential keywords, linguistic rules perform better \citep{hasan2014automatic}. However, it would be misleading to suggest that the survey responses are merely keywords surrounded by stopwords. The higher performance of our pipeline compared to a simple approach such as TF-IDF shows that there is added value in the extra steps involved in the pipeline.

Finally, the context provided by the survey itself makes a linguistic-based approach attractive. Unlike many other classification tasks, the answers are constrained into a topic (the survey question), which further restricts the number of potential keywords.

\section{Conclusion}

Initially, we set out some desirability criteria for an open source, computational pipeline for free form text data:

\begin{itemize}
\item Extensive reuse of existing models where possible;
\item Focus on relatively user-friendly pipelines, where the focus is on outputs that aid understanding of the data;	
\item Open-source code, in a high-level language such as Python;
\item Practical usability combined with transparency.
\end{itemize}

As sections 4 and 5 showed, the pipeline matches these criteria. Section 2 provided the basis for the importance of these criteria. In the past the analysis of large amounts qualitative, free-text research data has proven very time intensive and was always at risk of research bias leading to unreliable results. While there have been software packages that support the coding of responses to open ended questions, this still meant that the researcher had to read each and every response. With the advent of commercial text analytics packages for annotation this effort has been lessened, but still requires significant time and manual effort from researchers to extract keywords and topics. With purely statistical approaches to keyword extraction, including commercial software packages, the lack of transparency presents the researcher with a hidden and unquantified bias. In sum, these factors in many cases make free-text data unattractive for researchers when compared with ease with which large amounts of quantitative data can be analysed.

The software developed for this paper demonstrates how transparent, linguistically motivated keyword extraction can be performed in a way that allows for automating the analysis of large amounts of responses to open-ended questions in a survey. With an open source approach it is possible for an observer to test the process for possible biases in the way keywords are extracted. This will hopefully make it more feasible and attractive to apply qualitative questionnaires to much larger sample sizes. A side effect of larger sample sizes is that it allows for statistical analysis of qualitative data that also satisfies the requirements of quantitative researchers. Furthermore, due to the reduction in the time required to analyse large amounts of qualitative data, the automation of the analysis leads to a significant reduction in costs, making it viable to conduct qualitative research on a large scale for commercial purposes.

The profound consequences of this have been outlined at the beginning of this article. While the academic case for this has been made, we contend that ultimately the easiest way to make the case for more qualitative research is by showing a way to do the analysis of qualitative data in a quick and transparent way, producing appropriate summaries of the topics covered in large amounts of qualitative responses and let the results and ease of use speak for themselves.
Meanwhile, the open source Python implementation ensures that the pipeline is both flexible and available for extensions, modifications, and improvements

\subsection*{Acknowledgements}
This work was supported by The Alan Turing Institute under the EPSRC grant EP/N510129/1.

\subsection*{Author contributions}

All authors  contributed to the conceptualization of the study and for writing the original draft. 
BMcG and GBJ were responsible for developing the software and the computational methodology, and reviewing and editing the article. BMcG was additionally responsible for the evaluation, the statistical analysis, and the visualization. DH was responsible for acquiring the data. DH was responsible for the philosophical and practical justification for the software. All authors contributed to curating the data. All authors gave final approval for publication.

%for writing sections 1, 3, 4, 5,
 %GJ was responsible for writing sections 6 and 7. 
  %, and for writing section 2. 

%, X.X. and Y.Y.; methodology, X.X.; software, X.X.; validation, X.X., Y.Y. and Z.Z.; statistical analysis, Barbara McGillivray; resources, X.X.; data curation, X.X.; writing—original draft preparation, B + G + D; writing—review and editing, B and G; visualization, X.X..”
%- conceptualization B+G+D
%- software B+G
%- methodology B+G
%- evaluation B 
%- statistical analysis B 
%- resources (data) - D
%- manual analysis/data curation B+G+D not just D?
%- drafting of sections 
%    1 B+G??
%    2 D
%    3 B+G??
%    4 B+G
%    5 B+G??
%    6 G
%    7 G?
%- review and editing B+G
%- visualization B

% For research articles with several authors, a short paragraph specifying their individual contributions must be provided. The following statements should be used “conceptualization, X.X. and Y.Y.; methodology, X.X.; software, X.X.; validation, X.X., Y.Y. and Z.Z.; formal analysis, X.X.; investigation, X.X.; resources, X.X.; data curation, X.X.; writing—original draft preparation, X.X.; writing—review and editing, X.X.; visualization, X.X.; supervision, X.X.; project administration, X.X.; funding acquisition, Y.Y.”, please turn to the  \href{http://img.mdpi.org/data/contributor-role-instruction.pdf}{CRediT taxonomy} for the term explanation. Authorship must be limited to those who have contributed substantially to the work reported.

%\bibliographystyle{chicago}

\bibliography{bibliography}
\bibliographystyle{plainnat}

\end{document}